\newif\ifabstract
\abstracttrue
 \abstractfalse 
\newif\iffull
\ifabstract \fullfalse \else \fulltrue \fi

\documentclass[11pt]{article}
\usepackage{amsfonts}
\usepackage{amssymb}
\usepackage{amstext}
\usepackage{amsmath}
\usepackage{xspace}
\usepackage{theorem}
\usepackage{graphicx}
\usepackage{url}
\usepackage{graphics}
\usepackage{colordvi}
\usepackage{colordvi}
\usepackage{subfigure}
\usepackage{times}
\usepackage{graphicx} 
\usepackage{subfigure} 
\usepackage{array}
\usepackage{algorithm}
\usepackage{algpseudocode}
\usepackage{amsmath}
\usepackage{url}
\usepackage{color}
\input{./Definitions.tex}

\advance \oddsidemargin by -0.8in
\newcommand{\myparskip}{3pt}
\parskip \myparskip

        {\hspace*{\fill}$\Box$\par\vspace{4mm}}

\newcommand{\be}{\begin{enumerate}}
\newcommand{\ee}{\end{enumerate}}
\newcommand{\ed}{\end{description}}
\newcommand{\bi}{\begin{itemize}}
\newcommand{\ei}{\end{itemize}}

\def\square{\vbox{\hrule height.2pt\hbox{\vrule width.2pt height5pt \kern5pt
\vrule width.2pt} \hrule height.2pt}}

\renewcommand{\phi}{\varphi}

\mathchardef\hyphen="2D

\begin{document}

\title{Earliness-Aware Deep Convolutional Networks for \\Early Time Series Classification}
\author{Wenlin Wang\thanks{Department of Electrical \& Computer Engineering, Duke University, USA} \and Changyou Chen$^\ast$ \and Wenqi Wang\thanks{Department of Industrial Engineering, Purdue University, USA} \and Piyush Rai\thanks{Department of Computer Science \& Engineering, IIT Kanpur, India} \and Lawrence Carin$^\ast$}	


\maketitle

\thispagestyle{empty}
\begin{abstract}
We present Earliness-Aware Deep Convolutional Networks (EA-ConvNets), an end-to-end deep learning framework, for early classification of time series data. Unlike most existing methods for early classification of time series data, that are designed to solve this problem under the assumption of the availability of a good set of pre-defined (often hand-crafted) features, our framework can jointly perform feature learning (by learning a deep hierarchy of \emph{shapelets} capturing the salient characteristics in each time series), along with a dynamic truncation model to help our deep feature learning architecture focus on the early parts of each time series. Consequently, our framework is able to make highly reliable early predictions, outperforming various state-of-the-art methods for early time series classification, while also being competitive when compared to the state-of-the-art time series classification algorithms that work with \emph{fully observed} time series data. To the best of our knowledge, the proposed framework is the first to perform data-driven (deep) feature learning in the context of early classification of time series data. We perform a comprehensive set of experiments, on several benchmark data sets, which demonstrate that our method yields significantly better predictions than various state-of-the-art methods designed for early time series classification. In addition to obtaining high accuracies, our experiments also show that the learned deep shapelets based features are also highly interpretable and can help gain better understanding of the underlying characteristics of time series data.
\end{abstract}


\section{Introduction}

Early time series classification (ETSC), the task of predicting the label of a time series as early as possible, by only looking at an initial subsequence of the entire series, is becoming increasingly important in many time-sensitive domains. Examples include medical informatics and clinical prediction, weather reporting, financial markets, network traffic monitoring, etc~\cite{ghalwash2013extraction,xing2010brief,dainotti2011early,mao2012integrated,ghalwash2012early,mcgovern2011identifying}. While the problem of classifying \emph{fully observed} time series has been studied extensively in recent years, the earliness constraints posed by ETSC, which demands reliable \emph{early} predictions, makes it a considerably more challenging problem than standard time series classification. An ETSC system must ensure that it not only cuts down on the prediction time, but also makes \emph{reliable} predictions based on very few initial timestamps. These two goals are often conflicting with each other. 

In order to ensure reliable predictions from an ETSC system, it is desirable to have methods that can extract highly discriminative features, even from very limited observations in the time series. Unfortunately, most existing methods for this problem are usually based on ad-hoc ways of defining subsequences or hand-crafted patterns~\cite{ye2009time}, mapping them to fixed sized vectors, and subsequently applying a classification method (e.g., nearest neighbors or support vector machine). One of the key limitations of such approaches is that the extracted features can usually capture only label-based local patterns, and may have difficulties in generalizing to more complex time series data sets. Therefore, the ability to automatically learn highly discriminative features from (limited-length) time series data can be very important in these problems. Motivated by this, we present a deep learning framework for learning highly discriminative \emph{and} interpretable features in the context of early time series classification.

Shapelets~\cite{ye2009time} based representations have recently emerged as an effective way to model time series data in problems such as time series classification. Shapelets are subsequences in time series with good discriminative power. Shapelets also provide good interpretability by capturing the salient characteristics in the time series data. However, most of the existing methods perform shapelet extraction using time-consuming and ad-hoc ways. e.g., by trying out all possible prefix subsequences and evaluating them based on their ability to predict the target variable of interest. Some recent work has tried learning shapelets from data~\cite{grabocka2014learning} but the learned shapelets are usually not ideal for \emph{early} time series classification.

In this paper, we present a deep feature learning and classification framework based on an \emph{earliness-aware} deep convolutional neural network architecture. Our framework, henceforth referred to as Earliness-Aware Deep Convolutional Networks (EA-ConvNets), provides, to the best of our knowledge, the first deep, nonlinear feature learning based method for fast shapelet discovery, and integrates it with a nonlinear classification model to perform end-to-end deep learning for early classification of time-series data. Our deep learning framework can learn highly discriminative shapelets that are also highly interpretable, a property that is especially desirable in many ETSC problems, such as applications in medical and health informatics. As compared to the state-of-the-art methods for that can learn interpretable features for ETSC problems~\cite{xing2011extracting}, the deep shapelet based representations learned using our framework, in a purely data-driven manner, can capture patterns in the time series at multiple levels of granularities. 

Our framework is flexible to dynamic lengths of the input time series, and can make predictions at any time based on the current observations. A key step towards this goal is to train the ConvNet on \emph{stochastic truncated} time series data. Through an extensive set of experiments, we show that EA-ConvNets outperforms several state-of-the-art models for early time series classification, and is also competitive with time series classification methods that work with full time series.

\section{Convolutional Neural Nets for Time-Series}\label{sec:cnn}

Our framework is based on a convolutional neural network (CNN) architecture for time series data. We first provide a brief review of a basic CNN for time series data~\cite{yang2015deep}, before describing the details of our proposed framework. 

Consider a sequence representing a univariate time series $d=[ d_1, d_2, ..., d_L]$, where $L$ is the length of the sequence. A classical CNN for such time series data works by composing alternating layers of convolution and pooling operations. The convolutional layer extracts patterns within local regions throughout the input sequence. This operation is performed by convolving a filter, $g=[g_1, g_2, ..., g_m]$, of length $m$ over the sequence, and computing the inner product of the filter at each location in the sequence. For each layer, let us denote the outputs of this filter applied over the entire sequence, as a feature map vector \textbf{o}, which the $i$-th element written as $$\mathbf{o}_i=\sum_{j=1}^n g(j)\cdot d(i-j+\frac{L}{2})~.$$. 
Intuitively, it measures the similarity between the filter to each portion of the sequence. Afterwards, a nonlinear function $f(\cdot)$ is applied element-wise to the feature map $\mathbf{o}$ as $\mathbf{a}=f(\mathbf{o})$. The activations $\mathbf{a}$ are then passed to a pooling layer for nonlinear dimension reduction. 

The pooling layer aggregates the information within a set of small regions of $\mathbf{a}$, say $\{\mathbf{r}_k\}_{k=1}^K$ with $\mathbf{r}_k$ being some element of $\mathbf{a}$, and produces a \emph{pooled feature map} $\mathbf{s} = [\mathbf{s}_1,\ldots, \mathbf{s}_K]$ as the output. Let us denote the aggregation function to be $\textsc{pool}()$. Then for each feature map, $\mathbf{o}$, we have $$\mathbf{s}_k = \textsc{pool}(f(\{\mathbf{o}_i\}_{i \in \mathbf{r}_k}))$$ where $\mathbf{r}_k$ is pooling region $k$ in the feature map $\mathbf{o}$.


\section{Earliness-Aware ConvNets} \label{sec:EACon}

In this section, we formally define the early time series classification (ETSC) problem, and describe our framework, Earliness-Aware Deep Convolutional Networks (EA-ConvNets), in detail.

\subsection{Problem Definition}

We will denote a univariate time series (a single sequence) as $\Tb = \{ t_1, t_2, ..., t_L\}$, where $t_i$ is a scalar measurement\footnote{Extending to the vector-value setting is straightforward in our framework.} of this time series at timestamp $i$, $L$ is the total number of timestamps for this time series. Following this notation, we denote a collection of time series as $\Db=\{ (\Tb_i, y_i) \}_{i=1}^N$, where $N$ is the number of time series in the dataset, $\Tb_i = \{t_i^1, t_i^2, ... , t_i^L\}$ represents the $i$th time series and $y_i$ is the associated label. The element $t_i^j$ indicates the $j$th timestamp value in the $i$th time series. For simplicity of exposition, we will assume all the time series in $\Db$ to have identical timestamps, though our framework does not require this. 

The task in time series classification (TSC) is to build a classifier to predict the class label $y_*$ of a \emph{new} time series $\textbf{T}_*$, given past training data $\Db$. The time series in the training set as well as in the test data are assumed to be \emph{fully observed}.

In contrast, ETSC requires making prediction when the time series may only be \emph{partially} observed (i.e., upto a certain timestamp $l < L$) in the training and/or the test data. In these settings, it is desirable to predict class labels as early as possible, while retaining an acceptable level of accuracy.

\subsection{Earliness-Aware ConvNets}

We present a deep feature learning framework for early time series classification by building a multi-scale convolutional neural network (CNN) architecture, which is coupled with a time series truncation model that allows focusing on early parts of each time series. The multi-scale CNN enables extracting features at multiple scales from the time series data. In the context of time series data, the extracted features correspond to \emph{shapelets}~\cite{ye2009time} which are subsequences with high discriminative power. Unlike most other existing methods that extract shapelets from time series data using hand-crafted rules (which is often ad-hoc, tedious, and time-consuming), a distinguishing aspect of our framework is that the shapelets are learned in a purely data-driven manner. Although some recent methods have looked at ways of \emph{learning} shapelets that are optimized for a given data set~\cite{grabocka2014learning}, unlike our approach which learns a deep representation of the shapelets, these methods can only learn a shallow representation, which consequently are less expressive and have less discriminative power.   

\begin{figure*}[!htbp]
	\centering
	\includegraphics[trim={0cm 0cm 0cm 0cm}, clip, scale=0.4]{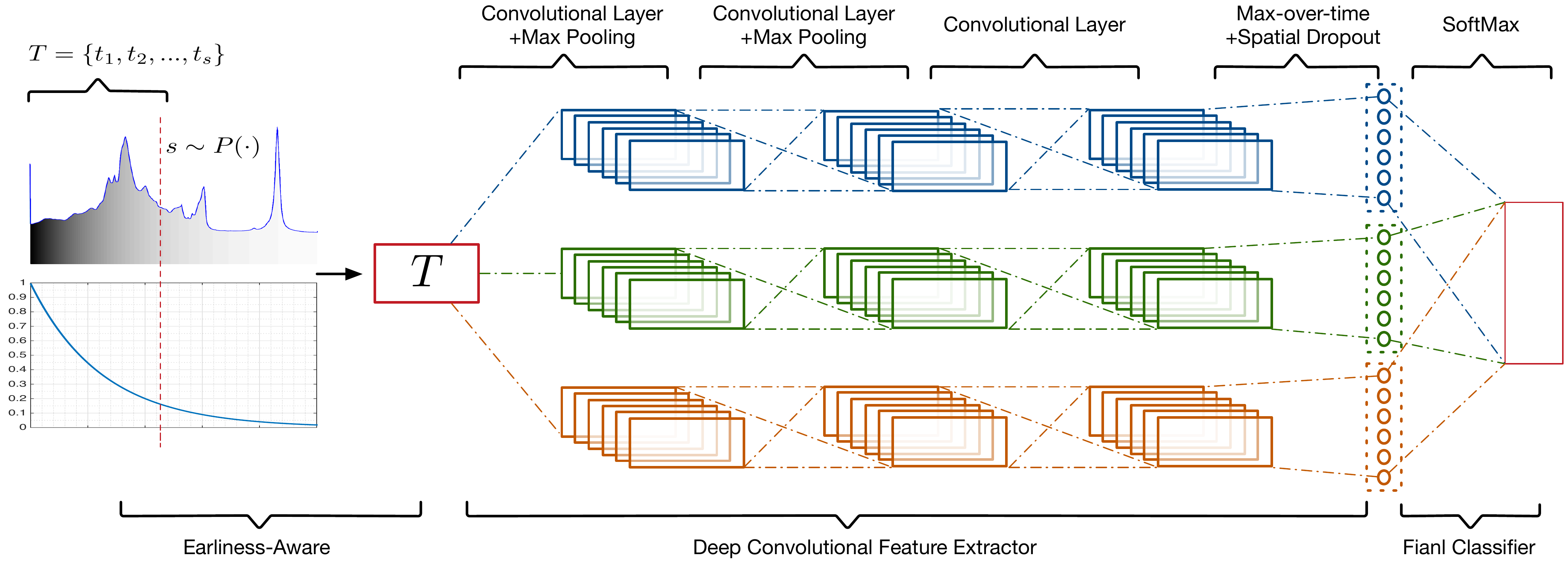}
	\caption{Overall achitecture of EA-ConvNets. In the earliness-aware part, the color of shade indicats the density of timestamps to be observed during training, and the figure below shows the exact probability distribution. }
	\label{Figure:Architecture}
\end{figure*}

We call our propose framework Earliness-Aware Deep Convolutional Networks (EA-ConvNets). The EA-ConvNets takes as input a raw time series and outputs the class label of that time series. The overall architecture of EA-ConvNets is shown in Figure~\ref{Figure:Architecture}, which consists of three stages: Early-Awareness, deep convolutional feature (shapelet) extractor, and the final classifier, each of which is described below.

\subsubsection{\textbf{Early-Awareness}} 
The Early-Awareness component lets EA-ConvNets to focus on the early stage of each time series. To accomplish Earliness-Aware, we draw inspiration from the idea of \emph{nested dropout}~\cite{rippel2014learning}, which was originally proposed to stochastically drop coherent nested sets of hidden units in a neural network by drawing a decay point from some decay distribution. We leverage a similar strategy to stochastically choose a truncation point for each time series, which gives rise to a new truncated version of the original time series (with the same label). To select the truncation point, a prior distribution $P(\cdot)$ on the timestamps (length) of the time series is employed. For each time series $\Tb_i$, we first sample an index $s\sim P(\cdot)$, generate the \emph{stochastically} truncated time series as ${\Tb_i}_{\downarrow s} = \{t_i^1, \cdots, t_i^s\}$, which then forms the input for the subsequent convolutional neural networks based deep shaplet detector module. For choosing the truncation point, as also done in~\cite{rippel2014learning}, we use a \textit{geometric distribution} because of its properties of exponential decay and memorylessness. The geometric distribution is parametrized as
\begin{align*}
	P(s; \rho) = \rho^{s-1}(1-\rho)~.
\end{align*}
An illustration of applying a geometric distribution based truncation of the original full-length time series is shown in the left sub-figure in Fig~\ref{Figure:Architecture}. Specifically, we refer to the process of drawing a truncation point $s$ from $P(\cdot)$, and truncating the time series at $s$ as \textit{s-truncation}; the \textit{s-truncation} of a time series $T$ is denoted as $T_{\downarrow s}$.


\subsubsection{\textbf{Deep Convolutional Feature (Shapelet) Extractor}}
The deep feature extraction stage of our framework is based on a deep convolutional neural network architecture, which takes as input the truncated time series obtained from the Earliness-Aware stage, extracts a hierarchy of shapelet based features for each time series, and passes these on to the final classifier. 

In order to extract the deep, shapelet based features, we apply independent $1D$ convolutions over the (truncated) time series. Since the time series data may exhibit latent features at multiple scales, a single filter of fixed length may not be able to capture this. Therefore, to capture the \emph{multi-scale} characteristics of the time series, we apply three channels of ConvNets in parallel. Each channel of the ConvNets has a different filter size, which is fixed within each ConvNet. 

In our experiments, we found that choosing the filter sizes from $\{3\%, 5\%, 10\%\}$ of $L$ performs well. The specific architecture is shown in Fig~\ref{Figure:Architecture}, with different filters shown in different colors. Applying each filter on the (truncated) time series gives as output a sequence of real values, each of which can be viewed as the similarity of a local subsequence of the time series with the applied filters. A larger filter size means a wider \textit{receptive field}, capturing a larger scale of information. 

After the convolutional operation, down-sampling is applied to reduce the output dimensions. Two types of down-sampling strategies are commonly used: average-pooling and max-pooling. In our experiments, we found that max-pooling produces better results and also leads to faster training. As a result, we adopt the max-pooling without overlapping. In addition to reducing the output dimension, down-sampling also has other benefits, such as capturing scale-invariant features and reducing model complexity to avoid overfitting.

We would also like to note that our EA-ConvNets framework is adaptive to dynamic timestamps. Denoting the output feature map of the last convolutional layer to be $O(\textbf{T}) = \{O_1(\textbf{T}),O_2(\textbf{T}), ..., O_M(\textbf{T})\}$, where $M$ is the number of filters in the layer. Note that the feature map $O_i(\textbf{T})$ itself is vector-valued (representing another time-series in a deep feature space). From the convolutional property, in the output feature map vector, the one with the largest value contains the most significant information. Collecting all these maximum values from the feature maps $O(\textbf{T})$ constitutes the final layer shapelet feature representation, $F(\textbf{T})$, for the time-series $\Tb$. This is essentially the \textit{max-over-time} method described in \cite{zhang2015character}, with the $i$th element of $F(\textbf{T})$ defined as $$F_i(\textbf{T}) = \underset{j}{\max}O_i(\textbf{T})[j]$$
Our feature extractor can be viewed as a hierarchical feature extractor, aiming for extracting high level abstract features, which correspond to shapelets learned in a purely data-driven fashion. To maintain the independence of each filter and avoid overfitting, we employ SpatialDropout\footnote{Applying DropOut on the convolutional layers is also possible, but it typically increases the computation burden without significantly improving the results \cite{krizhevsky2012imagenet}, we thus did not consider this setting.} (originally proposed in~\cite{tompson2015efficient}) on $F(\textbf{T})$. Our experiments demonstrate the effectiveness of the regularization achieved by SpatialDropout.

\subsubsection{\textbf{The Final Classifier}}  
For the final classifier, we concatenate all the shapelet based features from the outputs of the three ConvNets (one for each filter), and feed it into the final output layer, followed by a softmax transformation. The output of EA-ConvNets is the predictive distribution of each possible label for the input time series. The objective function is to minimize the cross-entropy loss defined as $$\mathcal{L}(\Wb) = -\sum_{i=1}^{N} \log p_{y_i}(\Tb_i)$$
where $\Wb$ represents all the model parameters ({\it e.g.}, the weights of the deep network), $p_{y_i}(\Tb_i)$ is the probability of $\Tb_i$ having the true label $y_i$, which is easily calculated by forwarding the input $\Tb_i$ thought the network (line 6 in Algorithm~\ref{alg:EA-ConNets}). Since all the operations in the objective function are differentiable, all the model parameters can be optimized jointly through backpropagation. 

The full algorithm is shown in Algorithm~\ref{alg:EA-ConNets}.
In the algorithm, Forward($\mathbf{B_{*}})$ in line 8 denotes the standard feedforward pass of a deep neural network, which takes a minibatch as input, and returns the loss on the corresponding minibatch. Backprop($L, y$) in line 9 is the standard backpropagation pass, which takes the loss and labels of a minibatch as input, calculates the gradients of model parameters, and then update the parameters via a stochastic optimization algorithm, {\it e.g.}, stochastic gradient descent. For the cross-validation in line 10, we propose a method to emphasize early-stage classification accuracy, with details provided in Section~\ref{sec:exp_setting} in the experiments.

\begin{algorithm}[tb]
	\caption{EA-ConvNets in pseudo-code.}
	\label{alg:EA-ConNets}
	\begin{algorithmic}[1]
		\State {\bfseries Input:} $\mathbf{D}=\{\Tb_i, y_i\}_{i=1}^N$; parameter of $P(\cdot)$ to be $\mathbf{\rho}$
		\State {\bfseries Output:} weights $\mathbf{W}$ of EA-ConvNets
		\State Initialize weights $\mathbf{W}$ of EA-ConvNets with uniform distribution range from $[-0.5, 0.5]$
		\For{\texttt{iteration}}
			\State Sample mini-batch $\mathbf{B}=\{\Tb_j, y_j\}_{j=1}^n$ from $\mathbf{D}$ 
			\State Sample a set of truncation variables $\mathbf{s} = \{s_1, \cdots, s_n\}$ for the 
				current mini-batch, where $s_j \sim P(\rho)$ 
			\State Get the \textit{s-truncation} mini-batch $\mathbf{B_{*}}=\{T_{\downarrow s_jj}, y_j\}_{j=1}^n$
			\State $L=\mbox{Forward}(\mathbf{B_{*}})$
			\State Backprop($L, \{y_j\}_{j=1}^n$)
			\State Evaluate on a validation set with EA-ConvNets
			\If{early-stop criteria satisfied}
				\State Break;
			\EndIf
		\EndFor
		\State Return $\mathbf{W}$
	\end{algorithmic}
\end{algorithm}

\section{Related Work}\label{sec:related}


Here we review the related work in two directions relevant to this work: algorithms for early time series classification (ETSC) and deep learning methods for time series classification (TSC). ETSC was arguably first proposed in \cite{de2004boosting} and developed along several independent lines, where classifiers make predictions from partial observations of temporal data. 

Some of the early work on ETSC is based on ensemble learning methods. In particular, \cite{mori2006early,uchida2008early} use boosting to facilitate a chronological restriction of features to achieve early recognition. More recently, \cite{ando2013time} proposed an ensemble of classifiers learned on subsequences of different lengths, and the prediction is made by the earliest individual classifier. Among other recent work, \cite{ando2016minimizing} learns the ensemble based early classifier by minimizing an empirical risk function and the response time required to achieve the minimum risk, which is formulized as a quadratic programming problem and solved by an iterative constraint generation algorithm.

Another prominent line of work is based on instance-based learning methods. In particular, \cite{xing2012early} proposed ETCS (Early Classification on Time Series), which introduced the notion of Minimum Prediction Length (MPL), the earliest timestamp for each time series in the training data to find the correct nearest neighbor, and searches all MPLs in a hierarchical fashion. At test time, for a given time series, we predict its label as that of its nearest neighbor with MPL. 

Inspired by the idea of shaplets~\cite{ye2009time}, which are subsequences of time series that can be highly discriminative for identifying a class label,  \cite{xing2011extracting}\cite{ghalwash2014utilizing}\cite{he2015early} extend instance-based methods by discovering a set of discriminative shapelets early in time. These methods predict a label for a test instance when a matching time series is found in the shaplets library.  Further, \cite{lin2015reliable} generalizes these methods to the multivariate setting with both numerical and categorical features. Another recent interesting approach is MD-MDPP+TD~\cite{li2014early}, which models multivariate time series as a Multivariate Marked Point-Process (Multi-MPP) and leverages the sequential cues as the temporal patterns to make early classification.  

Another class of methods for ETCS are based on learning a classifier and a detector simultaneously. The former aims to make real-time classifications while the latter is used to decide which result can be trusted and the label will be assigned subsequently. For example, \cite{ghalwash2012early} combines hidden markov models (HMM) and support vector machines (SVM) for early classification and set up a threshold on the prediction probability to ensure reliability. \cite{parrish2013classifying} proposes a local quadratic discriminant analysis (QDA) based approach. Its reliability is guaranteed by the threshold on the similarity of prediction on truncated and complete time series. \cite{dachraoui2015early} introduces a formal criterion for the detector to express the trade-off between earliness and accuracy. 

Among other recent work, \cite{weber2014lstm} utilizes the structure characteristic of the long short term memory (LSTM) to do early recognition during testing time. More recently, \cite{mori2016reliable} proposes a probability based classifier, which learns timestamps in which the prediction accuracy for each class surpass a user defined threshold. At test time, a decision will be made no early than the learned timestamps and reliability is controlled by the differences between the top two probabilities. 

For TSC, shaplets based methods have received a significant attention recently~\cite{grabocka2014learning}\cite{shah2016learning}\cite{kwok2016efficient}. Since, shaplets can be viewed as a special case of features learned by ConvNets~\cite{cui2016multi}, ConvNets to TSC is a promosing direction. Recently, \cite{zheng2014time} explored multi-channel CNN to deal with multivariate time series. Features are extracted through a parallel channel and then combined via an ensemble. In another recent work, \cite{cui2016multi} proposed a generalized CNN capable of extracting multi-scale characteristics in time series data and achieves state-of-the-art performance on multiple benchmark datasets. In contrast, ConvNets based models for early classification of time series have not been explored before and the EA-ConvNets framework proposed in our work is a first attempt in this direction.

\section{Experiments}\label{sec:exp}

We evaluate the effectiveness of EA-ConvNets on $12$ publicly available benchmark data sets. In the following, we give a brief description of the data sets and the experimental settings, and then conduct the following experiments
\begin{itemize}
 \item Evaluating our model for the task of early time series classification by comparing with several state-of-the-art methods for this problem.
 \item Comparing EA-ConvNets with state-of-the-art algorithms for {\em fully observed} time series data, and show that our model is also competitive with these algorithms in terms of classification accuracies.
 \item Visualizing the learned features (shapelets) in each of the layers of the deep architecture.
 \item Assessing the sensitivity of EA-ConvNets to the hyper-parameters.
 \item Comparing EA-Convnets with the setting of training multiple ConvNets, where each ConvNet is trained only on the data with a particular truncated level.
\end{itemize}

\subsection{Datasets}

We select $12$ benchmark data sets from the UCR time series archive \cite{UCRArchive}, with time series having a variety of lengths. Our data sets include: \textsc{Adiac},  \textsc{Fish}, \textsc{Gun Point}, \textsc{ItalyPowerDemand}, \textsc{Synthetic Control}, \textsc{Trace}, \textsc{Cricket-X}, \textsc{Cricket-Y}, \textsc{Cricket-Z}, \textsc{Two Patterns}, \textsc{Non-Invasive Fetal ECG Thorax1}(\textsc{NonInvThorax1}) and \textsc{Non-Invasive Fetal ECG Thorax2}(\textsc{NonInvThorax2}). Some statistics of the data sets are listed in Table~\ref{Table:data}.  Our experiments use the default training and test splits provided by UCR.

\begin{table}[!htbp]
	\setlength{\tabcolsep}{4pt}
	\caption{Summary of datasets used in the evaluation.}
	\label{Table:data}
	\begin{center}
		\begin{small}
			\begin{tabular}{l>{\centering}m{2cm}cccr}
				\hline
				Dataset  & $\# train$ & $\# test$ & $|Y|$ & $d$  \\
				\hline
				\textsc{Adiac}                     & $390$	 & $391$  & $37$ & $176$ \\
				\textsc{Fish}             		   & $175$	 & $175$  & $7$  & $463$ \\
				\textsc{Gun Point}         		   & $50$    & $150$  & $2$  & $150$ \\
				\textsc{ItalyPowerDemand}          & $67$    & $1029$ & $2$  & $24$  \\
				\textsc{Synthetic Control}         & $300$   & $300$  & $6$  & $60$  \\
				\textsc{Trace}                     & $100$   & $100$  & $4$  & $275$ \\
				\textsc{Cricket-X} 				   & $390$	 & $390$  & $12$ & $300$ \\
				\textsc{Cricket-Y} 				   & $390$	 & $390$  & $12$ & $300$ \\
				\textsc{Cricket-Z} 				   & $390$	 & $390$  & $12$ & $300$ \\
				\textsc{Two Patterns}              & $1000$  & $4000$ & $4$  & $128$ \\
				\textsc{NonInvThorax1}             & $1800$	 & $1965$ & $42$ & $750$ \\
				\textsc{NonInvThorax2}             & $1800$	 & $1965$ & $42$ & $750$ \\
				\hline
			\end{tabular}
		\end{small}
	\end{center}
\end{table}

\subsection{Experimental Setting}\label{sec:exp_setting}

For consistency, we conduct all the experiments on all the data sets using the same architecture as shown in Fig~\ref{Figure:Architecture}. The number of filters is set to $\{48, 48, 96\}$ for each layer in each channel of the networks,  and is the same across the three channel of the network. As a result, the final feature representation has a size $96\times 3 = 288$. ReLU is adopted as the nonlinear activation function after each hidden layer. EA-ConvNets is implemented using Torch7 \cite{collobert2011torch7} and trained on NVIDIA GTA TITAN graphics cards with 6GB RAM. We optimize the model with RMSprop \cite{TielemanH:12} using minibatches of size 50. We use 5-fold cross-validation for hyperparameter tunning.

Specifically, the hyperparameters of EA-ConvNets include the value of geometric distribution hyperparameter $\rho$, pooling factor chosen from among $\{2, 3, 5\}$, SpatialDropout rate chosen from among $\{0.4, 0.5, 0.6\}$ and $L_2$ weight decay chosen from $\{0, 10^{-6}, 10^{-5}, 10^{-4}\}$. We apply early stopping to avoid overfitting. We use ``AUC'' on the validation set as the criteria to determine when to early stop. The``AUC'' (not the usual area under the ROC curve) indicates the area under a curve, plotting  the trend of average truncation length vs accuracy. Early stopping is triggered when the maximum ``AUC'' on validation sets maintains for a number of epochs. 

Also note that, unlike several existing time series classification methods, we use raw time series as input and do not use any other data augmentation, except for the stochastic truncation. The automatic feature learning capability of our deep model allows it to learn the best features for the task.

 \subsection{Baselines}
 
We conduct two types of experiment: (1) early time series classification (ETSC), and (2) fully observed time series classification (TSC). For the former, we compare EA-ConvNets against the following state-of-the-art baselines:
 
 \begin{itemize}
 	\item Early classification on time series (ECTS)\cite{xing2012early}
 	\item Early Distinctive Shaplet Classification (EDSC) \cite{xing2011extracting}
 	\item Reliable early classification method (RelClass) \cite{parrish2013classifying}
	\item Reliable early classification framework for time series based on class discriminativeness and reliability of predictions (ECDIRE) \cite{mori2016reliable}
	\item Deep convolutional neural network classifier with the same network architecture as EA-ConvNets with \textit{full observation} as the training set (EA-ConvNets-Full). 
 \end{itemize}
 
Note that our baselines also include methods that can learn \emph{shapelets} from data (EDSC), just like EA-ConvNets (which however can learn a deep feature representation of the shapelets). 

The goal of the latter experiment is to show that EA-ConvNets is able to do early classification without compromising much on the classification accuracies as compared to methods that can work with fully observed time series. For this experiment, we compare EA-ConvNets against the following state-of-the-art baselines for fully-observed time series:

 \begin{itemize}
 	\item $1$-Nearest Neighbor ($1$NN) classifier under \textit{full observation} with Euclidean distance\cite{faloutsos1994fast}
 	\item $1$NN with Dynamic Time Warping (DTW) \cite{berndt1994using}
 	\item Bag-of-SFA-Symbols (BOSS) \cite{schafer2015boss}
 	\item Elastic Ensemble (PROP) \cite{lines2015time}
 	\item Learning Shapelets Models(LTS) \cite{grabocka2014learning}
 	\item COTE~\cite{bagnall2015time}, a weighted vote of over $35$ classifiers
 	\item Multi-scale convolutional nerual networks (MCNN)\cite{cui2016multi}
 \end{itemize}

\subsection{Early Time Series Classification}

\begin{figure*}[!htbp]
	\centering
	\includegraphics[trim={0cm 0cm 0cm 0cm}, clip, scale=0.6]{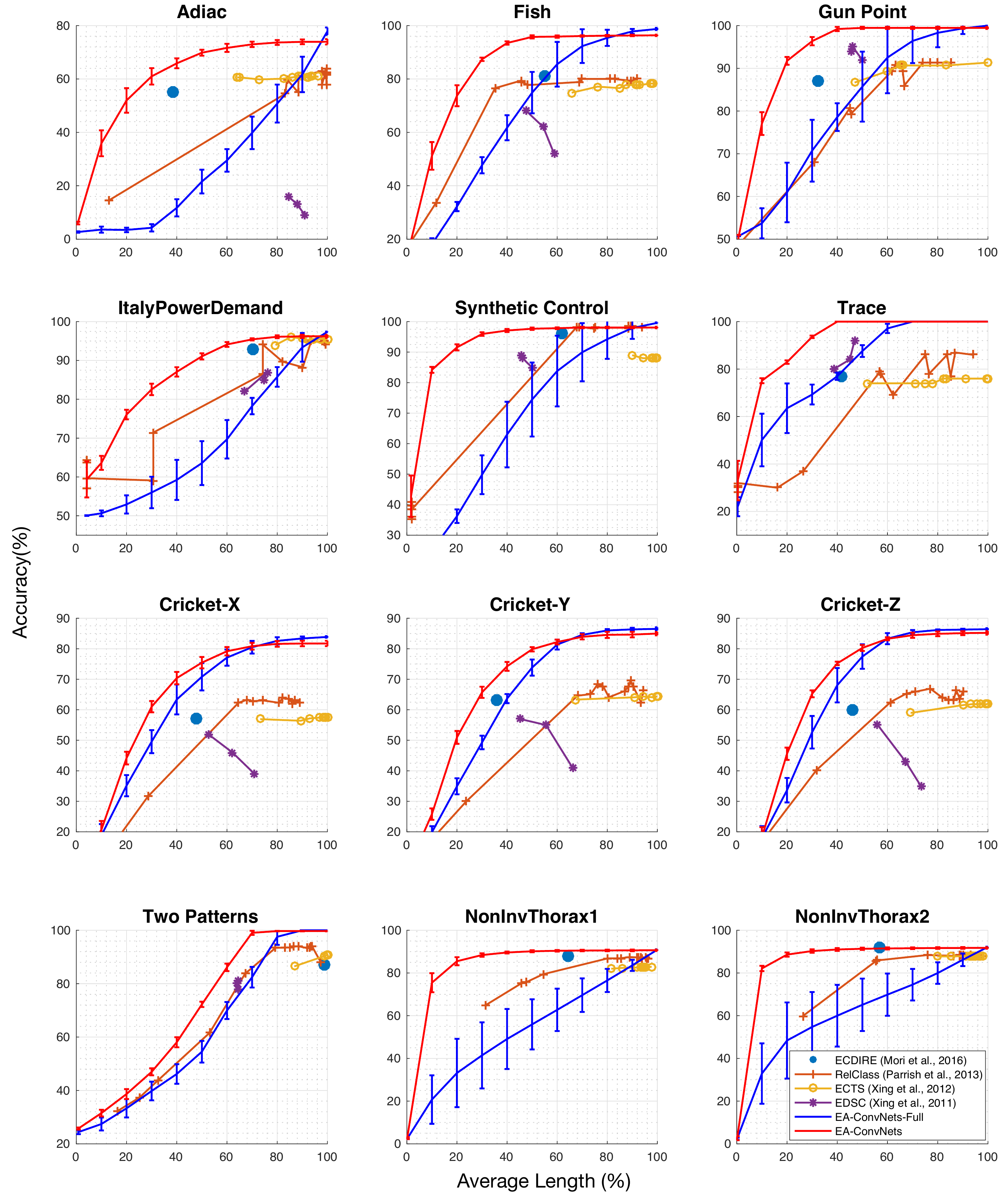}
	\caption{Test Accuracy with varying length of timestamps. The average length has been normalized to be the ratio to the full observation. }
	\label{Figure:AccVsLength}
\end{figure*}

In this section, we compare EA-ConvNets with several state-of-the-art baseline methods for early time series classification. To assess the robustness of all the methods on short time series data, for each data set, we also evaluate each method on varying lengths of the input time series (Fig~\ref{Figure:AccVsLength}).

For ECTS~\cite{xing2012early}, we set the parameter \textit{minimum support} from 0 to 1 with the dense interval to be 0.05. For EDSC~\cite{xing2011extracting}, we adopt the version based on Chebyshev Inequality and set the bound for Chebyshev condition to be $2.5$, $3$ and $3.5$, which has been reported to be efficient by the authors. For RelClass~\cite{parrish2013classifying}, we choose the values for the reliability threshold from among $\{10^{-30}, 10^{-25}, 10^{-20}, 10^{-15}, 10^{-10}, 10^{-5},10^{-4}, 10^{-3}, 10^{-2}, \\ 10^{-1}, 0.25, 0.50, 0.75, 0.90, 0.99\}$ and for each individual test, the reliability threshold is calculated with \textit{Gaussian Naive Bayes box} method. In contrast, for EA-ConvNets, a range of classification thresholds is set for the output of the SoftMax layer in EA-ConvNet. For a particular threshold, if the highest confidence score (the output of the SoftMax layer) surpasses the value, we accept the prediction for the current time series (thus can determine if the time series is classified correctly, and in the end calculate the average length under this threshold); otherwise we proceed until next available timestamps. The overall results are plotted in Fig~\ref{Figure:AccVsLength}, which show that. 

\begin{enumerate}
\item For all fractions of the time series given as input to each method, EA-ConvNets outperforms the other baselines on all the data sets by a significantly large margin, except for \textsc{NonInvThorax1} and \textsc{NonInvThorax2} data sets, on which the ECDIRE baseline is comparable to EA-ConvNets. This demonstrates the robustness and generality  of EA-ConvNets. 
\item Even when the full time series is given, EA-ConvNets is competitive with EA-ConvNets-Full (which does not perform stochastic truncation and thus does not have the early-awareness). This indicates that the earliness aspect of EA-ConvNets does not compromise on the classification accuracies. 
\item It is also worthwhile to note that the variance of EA-ConvNets is much smaller than that of EA-ConvNets-Full. Therefore, earliness-aware can also be viewed as a robust regularizor (akin to a ``dropout'' mechanism). 
\item EA-ConvNets is particularly effective at {\em extreamly early classification}. This is evidenced especially from the high classification accuracies even when only 20\%-40\% of the time series is given as the input to each of the baselines and EA-ConvNets. 
\end{enumerate}

\begin{table*}[!htbp]
	\setlength{\tabcolsep}{6pt}
	\caption{Test Errors: Comparison with (Full) Time-Series Classification Methods}
	\label{Table:Accuracy}
	\begin{center}
	\resizebox{\columnwidth}{!}{%
	\begin{tabular}{c | c   c  c  c c c c c c c c r}
		\hline
		Dataset & 1NN(ED) & DTW & BOSS  &PROP &LTS  & COTE & MCNN & EA-ConvNets-Full & EA-ConvNets \\
		\hline
		\textsc{Adiac} & $0.389$ & $0.391$ & $\mathbf{0.220}$ & $0.353$ &  $0.437$ & $0.233$ & $0.231$ & $0.221\pm 0.016$ & $0.252\pm 0.008$\\
		\textsc{Fish}  & $0.217$ & $0.160$ & $\mathbf{0.011}$ &  $0.034$ & $0.066$ &  $0.029$ & $0.051$ & $0.013\pm 0.005$ & $0.037\pm 0.003$\\
		\textsc{Gun Point} & $0.087$ & $0.087$ & $\mathbf{0}$ &   $0.007$ & $\mathbf{0}$ &  $0.007$ & $\mathbf{0}$ & $\mathbf{0\pm 0}$ & $0.005\pm 0.003$\\
		\textsc{ItalyPowerDemand} & $0.045$ & $0.045$ & $0.053$ &  $0.039$ & $0.030$ &  $0.036$ & $0.030$ & $\mathbf{0.023\pm 0.001}$ & $0.038\pm 0.005$ \\
		\textsc{Synthetic Control} & $0.120$ & $0.017$ & $0.030$ &  $0.010$ & $0.007$ &  $\mathbf{0}$ & $0.003$  & $0.004\pm 0.002$ & $0.019\pm 0.003$\\
		\textsc{Trace} & $0.240$ & $0.01$ & $\mathbf{0}$ &  $0.010$ & $\mathbf{0}$ &  $0.010$ & $\mathbf{0}$ & $\mathbf{0\pm 0}$ & $\mathbf{0\pm 0}$\\
		\textsc{Cricket-X}& $0.426$ & $0.236$ & $0.259$ &   $0.203$ & $0.209$ & $\mathbf{0.154}$ & $0.182$ & $0.161\pm 0.003$ & $0.183\pm 0.010$ \\
		\textsc{Cricket-Y} & $0.356$ & $0.197$ & $0.208$ &   $0.156$ & $0.249$ &  $0.167$ & $0.154$ & $\mathbf{0.135\pm 0.006}$ & $0.150\pm 0.006$\\
		\textsc{Cricket-Z} & $0.379$ & $0.180$ & $0.246$ &   $0.156$ & $0.201$ &  $\mathbf{0.128}$ & $0.142$ & $0.136\pm 0.004$ & $0.148\pm 0.007$\\
		\textsc{Two Patterns} & $0.093$ & $0.002$ & $0.016$ &   $0.067$ & $0.003$ &  $\mathbf{0}$ & $0.002$ & $\mathbf{0\pm 0}$ & $0.003\pm 0.001$\\
		\textsc{NonInvThorax1}& $0.171$ & $0.185$ & $0.161$ &   $0.178$ & $0.131$ &  $0.093$ & $\mathbf{0.064}$ & $0.091\pm 0.002$ & $0.093\pm 0.004$\\
		\textsc{NonInvThorax2}& $0.120$ & $0.129$ & $0.101$ &   $0.112$ & $0.089$ &  $0.073$ & $\mathbf{0.060}$ & $0.078\pm 0.001$ &$0.083\pm 0.004$\\
		\hline
	\end{tabular}
}
	\end{center}
\end{table*}
\vspace{-0.5em}
\subsection{Comparison with (Full) Time Series Classification Methods}
We also compare EA-ConvNets with several state-of-the-art baselines for full time series classification (TSC). The results are shown in Table~\ref{Table:Accuracy}. None of the previous work on has compared ETSC classifiers with TSC classifiers under full observations, due to the fact that ETSC trades off accuracies for early classification with full classification\footnote{As a result, ETSC usually performs worse than TSC under full observations.}. To the best of our knowledge, we are the first to conduct such experiments, to investigate the extent of the gap between ETSC and TSC algorithms. As can be seen in Table~\ref{Table:Accuracy}, EA-ConvNets with full observations outperforms most TSC baselines on multiple datasets, and is mostly competitive with state-of-the-art models. For some datasets, though EA-ConvNets does not outperform the state-of-the-art baselines, the gaps are marginal. We would like to stress that EA-ConvNets is not designed for full time-series classification tasks, though a comparable performance could be achieved as compared to state-of-the-art TSC baselines. Therefore, EA-ConvNets is practically a reliable model for anytime time-series classification.

\subsection{Qualitative Results: Deep Shapelet Discovery}
An appealing aspect of our framework is its ability to learn a deep feature representation of the time series data in form of a hierarchy of shapelets (highly discriminative subsequences) automatically from data. In addition to being highly discriminative features, shapelets can also be useful in understanding/visualizing the characteristics of the underlying time series and can be useful in other decision-making tasks (e.g., in medical diagnosis). To show the interpretability of the shapelets learned by EA-ConvNets, we visualize the learned shapelets at each stage of the deep architecture, using data set \textsc{Trace} as an example. 

\begin{figure*}[!htbp]
	\centering
	\includegraphics[trim={0cm 0cm 0cm 0cm}, clip, width=\textwidth]{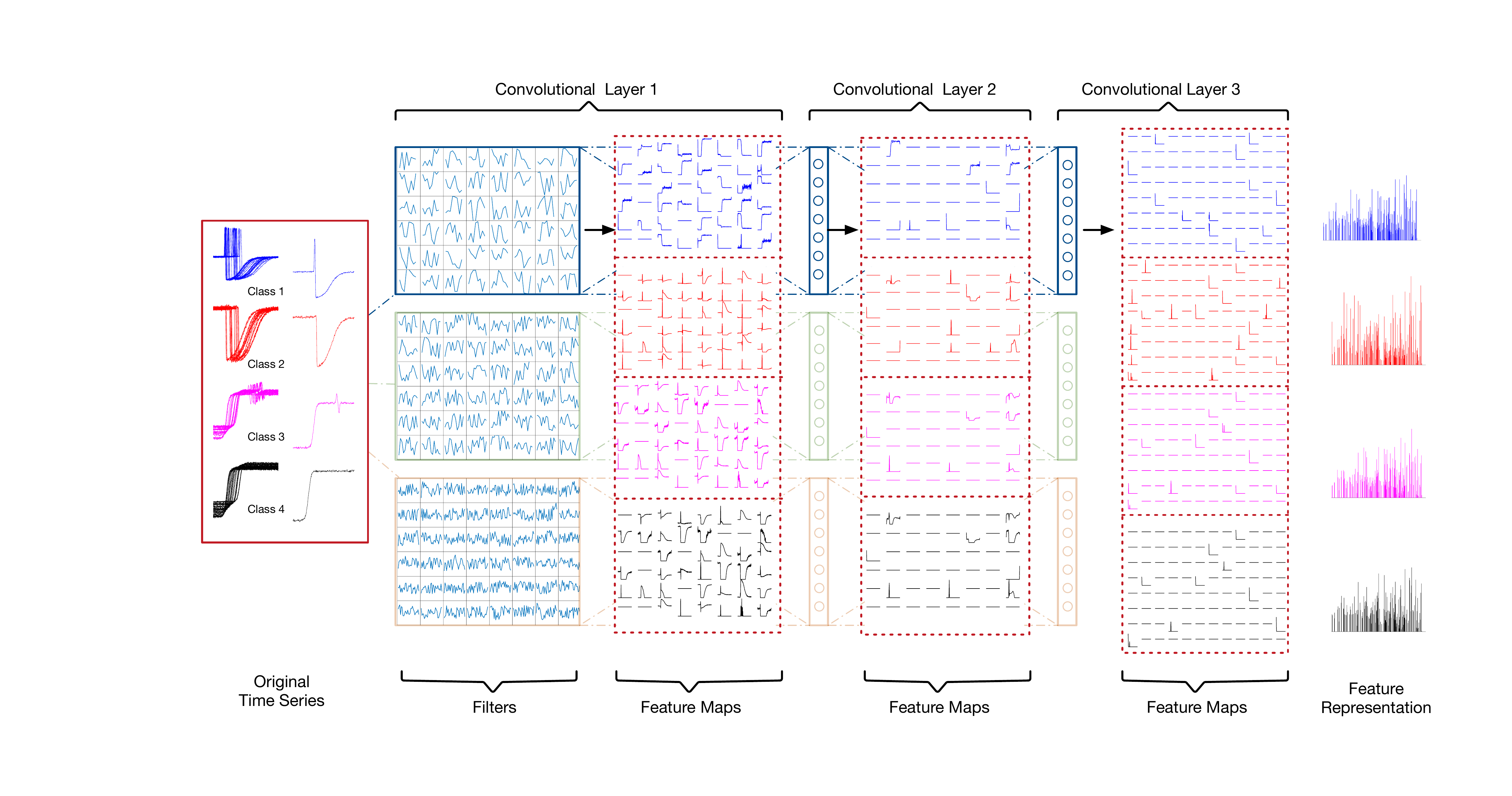}
	\caption{Visulization of dataset \textsc{Trace}. The plot in the left red rectangle is the visualization of the entire datasets and one sample from each class, which is fed to the EA-ConvNet for follow up visualization. Different colors indicate different categories. The plot inside the dotted red rectangle is the output feature maps after each correponding layers. The patterns shown in the first layer is the learned filters of EA-ConvNets. The histogram on the right is the feature representation after max-over-time. }
	\label{Figure:Visual}
\end{figure*}

We visualize the learned filters across all the three channels of the network. To visualize the feature maps of a given time series at each stage, we only plot those in the first channel for ease of visualization. Feature maps in the other two channels show similar patterns as the first channel, and are now shown here. Fig~\ref{Figure:Visual} shows the results. 

It is remarkable to note that our EA-ConvNets can automatically learn different layers of features (shapelets) simultaneously, as shown in the first layer of the networks. These features can be either low or high frequency patterns, or a mixture, depending on the data. By comparing the output feature maps of each layer, we notice that the deeper the layer goes, the more abstract features it can learn, which is typically a group of sparse and spiked features. It is also interesting to note that, for each feature map (either a sparse vector or a vector with spike in Figure~\ref{Figure:Visual}) in the final layer, the spike (corresponding to a spike in the final feature representation) usually appears at the beginning of the feature map. This indicates that the highly discriminative features are extracted in the early stages of a time series, which provides a qualitative evidence of the effectiveness of EA-ConvNets. Finally, as shown in the figure, the final feature representation for different classes distribute discriminatively. Therefore, EA-ConvNets is a powerful model to capture robust, highly interpretable, deep features (which correspond to a \emph{deep} hierarchy of shapelets) to make early, reliable predictions.

\begin{figure*}[!htbp]
	\centering
	\includegraphics[trim={0cm 0cm 0cm 0cm}, clip, width=\textwidth, height=4.5cm]{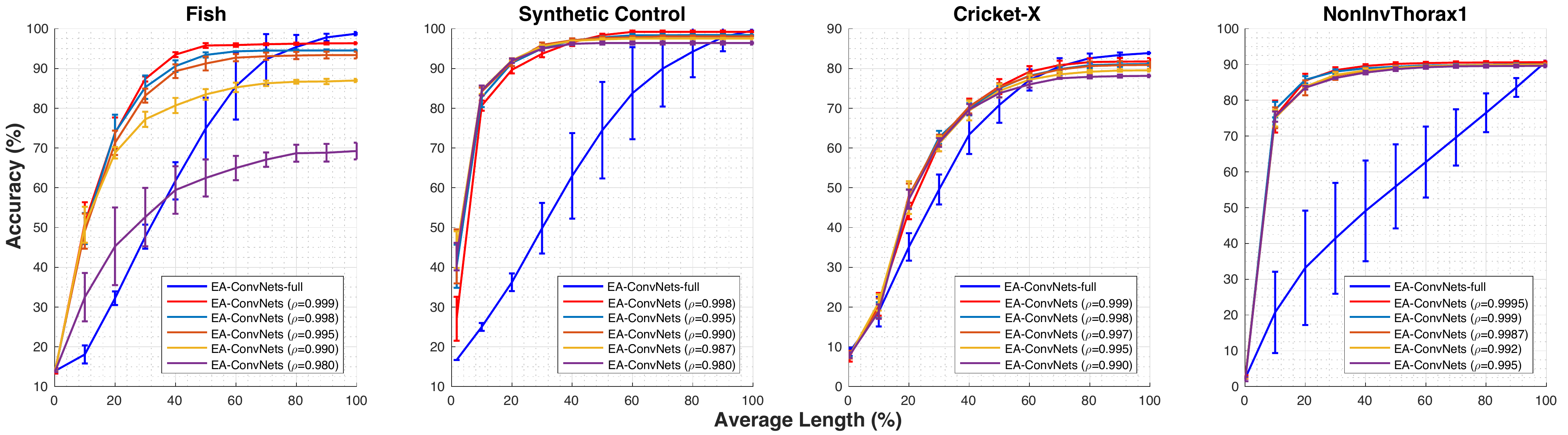}
	\caption{Results by varying the hyperparameter $\rho$}
	\label{Figure:Sensitivity2rho}
\end{figure*}  

\begin{figure*}[!htbp]
	\centering
	\includegraphics[trim={0cm 0cm 0cm 0cm}, clip, width=\textwidth, height=4.5cm]{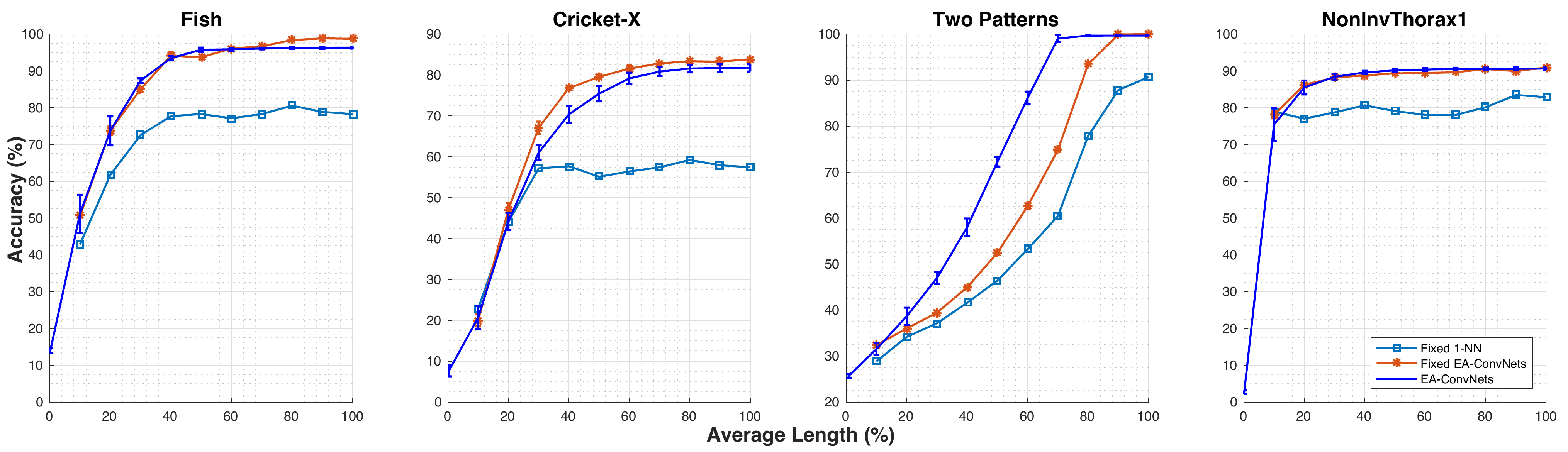}
	\caption{Comparison of EA-ConvNets with ensemble of models constructed using a variety of truncated levels}
	\label{Figure:MultiModel}
\end{figure*}

\subsection{Sensitivity to hyperparameters}

EA-ConvNets includes a numbers of hyperparameters. Among them, the pooling factor, SpatialDropout rate and $L_2$ weight decay, affect the convergence of the training algorithm. The hyperparameter, $\rho$, in the geometric distribution for data augmentation, not only affects the convergence of the training algorithm, but also controls the degree of earliness-aware of our model. Therefore, in this section, we mainly analyze the sensitivity of our model to $\rho$. We use \textsc{Fish}, \textsc{Synthetic Control}, \textsc{Cricket-X} and \textsc{NonInvThorax1} data sets to show the effect of this hyperparameter. The results of EA-ConvNets on these datasets for different values of $\rho$ are shown in Fig ~\ref{Figure:Sensitivity2rho}. As we can see, the effect of $\rho$ varies across different data sets. On \textsc{Fish}, the performance is sensitive to $\rho$. With an increase in $\rho$,  EA-ConvNets results in higher accuracies across the board for various fractions of the entire time series. The story is slightly different on the \textsc{Synthetic Control} and \textsc{Cricket-X} data sets, where with an increase on $\rho$, the curves converge to a higher accuracies under full observations but the accuracies are much lower when the fraction of observed time-series is small. This is because the larger $\rho$ is,  the higher the probability is that EA-ConvNets will observe most of the later time steps of the time series. As a result, the model would move its awareness towards the later parts of the time series. 
There is a trade-off between the accuracy and the earliness-aware when $\rho$ increases. On the other hand, on \textsc{NonInvThorax1}, for a wide range of values of $\rho$, the accuracies are roughly the same. This can be possibly because the highly discriminative features are present in the early stage for particular data set, and are captured effectively by EA-ConvNets. 
To summarize, although EA-ConvNets relies on a good choice of this hyperparameter, for a sufficiently broad range of value of $\rho$, our model yields reasonable results. Nevertheless, we recommend using cross validation to select an appropriate $\rho$ for different data sets.

\subsection{Comparison with Ensemble of Multiple Models}
To further evaluate the reliability of EA-ConvNets, we compare it with a combination (ensemble) of multiple models, each trained by truncating the original time series to a fixed length. Specifically, we compare EA-ConvNets with a group of deep learning models sharing the same architecture as EA-ConvNets but having a fixed truncation length (Fixed EA-ConvNets) and multiple $k$NN classifiers with fixed truncation length (Fixed $1$-NN). For each truncation length, we train independent models and use them at test time to predict the label of a test time series having the same truncation length. We conduct experiments on four data sets: \textsc{Fish}, \textsc{Cricket-X}, \textsc{Two Patterns} and \textsc{NonInvThorax1}. Fig~\ref{Figure:MultiModel} shows the results. 

It is interesting to see that on data set \textsc{Two Patterns}, our EA-ConvNest outperforms both Fixed EA-ConvNets and Fixed $1$-NN by a significant margin. The reasons may be attribute to: (1) the dynamic adaptation of EA-ConvNets. EA-ConvNets performs prediction adaptively, based on a threshold on prediction confidence; while for the ensemble of models, each model is restricted to predict based on a particular truncation length, irrespective to specific data sets; and  (2) the ability to deal with noise in the time series with EA-ConvNets. When the time series data contains noise, there is no guarantee for higher prediction accuracy with longer time series length. Our EA-ConvNets provides a proper way to deal with the noise by making early predictions with a reasonable prediction confidence. This is especially reflected in the results on \textsc{NonInvThorax1}, where EA-ConvNets performs consistently similar to Fixed EA-ConvNets, with the accuracies for both saturating after roughly 20\% of the observed time-series length. Note although EA-ConvNets is a little weaker than the corresponding multiple models on the \textsc{Cricket-X} dataset, the gap is very marginal.

\section{Conclusion} \label{sec:con}

We have introduced EA-ConvNets, a deep model based on a multi-scale convolutional neural network architecture for early classification of time series data. EA-ConvNets leverages the information at different scales and captures the interpretable features (shapelets) at a very early stages, which also makes it a robust model when each time series is inherently of a short length. Our experiments indicate that EA-ConvNets yields lower or comparable test errors given a pre-defined observed length budget at test time, while maintaining comparable classification accuracies to the state-of-the-art time series classification models that use full time series.
To interpret the effectiveness of EA-ConvNets, we present the learned local features (shapelets) and visualize the nonlinear feature representations in each layers of the deep model. Our experiments clearly show that the learned feature representation is highly discriminative and can achieve effective early classification. 

There are a number of interesting future directions. One direction is to extend our earliness-aware framework using other neural models such as LSTMs, which are natural tools for deal with time series data. Moreover, although in this paper we only considered EA-ConvNets for univariate time series, the framework can be extended to multivariate time series data sets. Finally, our framework is not only limited to classification tasks, and can be extended to more general (early) time series prediction tasks.

\bibliographystyle{plain}
\bibliography{references/EA_CNN}

\end{document}